\documentclass[runningheads]{llncs}

\usepackage[year=2024,ID=7380]{eccv}

\usepackage{eccvabbrv}

\usepackage{graphicx}
\usepackage{booktabs}
\usepackage{caption}
\usepackage[accsupp]{axessibility}  

\usepackage[pagebackref,breaklinks,colorlinks,citecolor=eccvblue]{hyperref}

\usepackage{orcidlink}
\usepackage{tabularx}
\usepackage{multirow}
\usepackage{array}
\usepackage{tcolorbox}
\newcolumntype{P}[1]{>{\centering\arraybackslash}p{#1}}
\begin{document}

\title{DeepClean: Machine Unlearning on the Cheap by Resetting Privacy Sensitive Weights using the Fisher Diagonal} 


\author{Jialei Shi \and
Najah Ghalyan \and
Kostis Gourgoulias \and
John Buford \and
Sean Moran
}

\authorrunning{J.~Shi et al.}

\institute{JP Morgan Chase \\
\email{\{jialei.shi, najah.ghalyan, john.buford, sean.j.moran}\}@jpmchase.com, \email{kostis.gourgoulias}@jpmorgan.com\\ }

\maketitle

\begin{abstract}
  Machine learning models trained on sensitive or private data can inadvertently memorize and leak that information. Machine unlearning seeks to retroactively remove such details from model weights to protect privacy. We contribute a lightweight unlearning algorithm that leverages the Fisher Information Matrix (FIM) for selective forgetting. Prior work in this area requires full retraining or large matrix inversions, which are computationally expensive. Our key insight is that the diagonal elements of the FIM, which measure the sensitivity of log-likelihood to changes in weights, contain sufficient information for effective forgetting. Specifically, we compute the FIM diagonal over two subsets -- the data to retain and forget -- for all trainable weights. This diagonal representation approximates the complete FIM while dramatically reducing computation. We then use it to selectively update weights to maximize forgetting of the sensitive subset while minimizing impact on the retained subset. Experiments show that our algorithm can successfully forget any randomly selected subsets of training data across neural network architectures. By leveraging the FIM diagonal, our approach provides an interpretable, lightweight, and efficient solution for machine unlearning with practical privacy benefits. Code release is depend upon the acceptance of paper.
\end{abstract}

\section{Introduction}
\label{sec:intro}

Machine learning models have achieved impressive results across many domains, but their continued deployment raises concerns around privacy, fairness, and model governance. Once a model has been trained on certain data, it can be challenging to fully ``unlearn'' that information. Models trained on problematic, biased, or private data can perpetuate harm even when deployed with good intentions. While collecting clean and ethical training data is ideal, it is not always feasible nor efficient to retrain models from scratch. Instead, we need methods to retroactively "unlearn" sensitive information from deployed models. Machine unlearning techniques aim to selectively remove information from trained models to address these concerns without requiring full retraining.

\begingroup
\captionsetup{belowskip=-20pt}
\begin{figure}[t]
\centering
\includegraphics[width=0.9\textwidth]{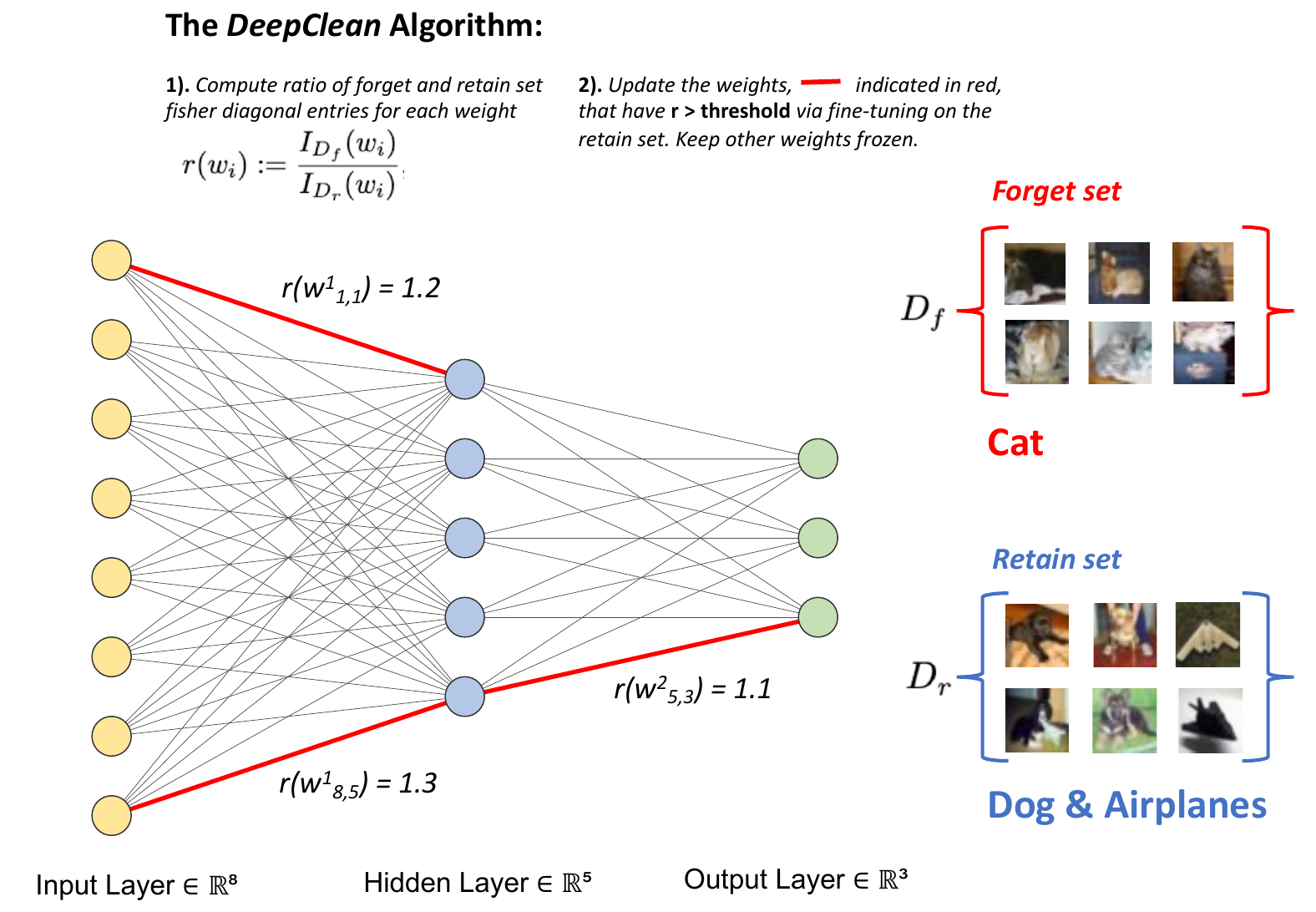}
\caption{Overview of DeepClean, a new baseline for machine unlearning. While conceptually simple, DeepClean is computationally efficient and empirically effective (see below).}
\label{fig:overview-deepclean-arch}
\end{figure}
\endgroup

A nascent but growing field of research has begun exploring techniques for machine unlearning - selectively removing sensitive knowledge already encoded in trained models. Proposed approaches include leveraging influence functions to identify and forget influential training examples~\cite{koh2017understanding}, approximating the Fisher Information Matrix to efficiently update parameters~\cite{liu2023unlearning}, imposing regularization that facilitates forgetting~\cite{afonja2022generative}, and using generative models to synthesize replacement data~\cite{Jinsung_2020}. While demonstrating promise, existing machine unlearning algorithms define unlearning differently. It is crucial to acknowledge a good machine unlearning algorithm should

\begin{itemize} 
\item Be model agnostic.
\item Not be aware of intermediate training stages, \textit{e.g.} no need to store gradient or model weights during training.
\item Not require special training paradigm when having the model to be unlearned, \textit{e.g.}~sharding training data or model.
\item Be able to remove either random or same label samples.
\end{itemize}
Our work, dubbed \emph{DeepClean} - illustrated in Fig.~\ref{fig:overview-deepclean-arch} - aims to advance the field by developing an efficient, interpretable and real-life usecase adaptable unlearning method, which adheres to the proposed rules by analyzing the diagonal Fisher Information Matrix across training subsets. This enables sensitive forgetting without requiring model retraining or full dataset access. We introduce an optimization framework to identify a small subset of weights that, if reset, maximally reduces information related to sensitive information in the data while minimally impacting accuracy on other attributes of the task.

\begin{figure}[h]
\centering
\includegraphics[width=1\textwidth]{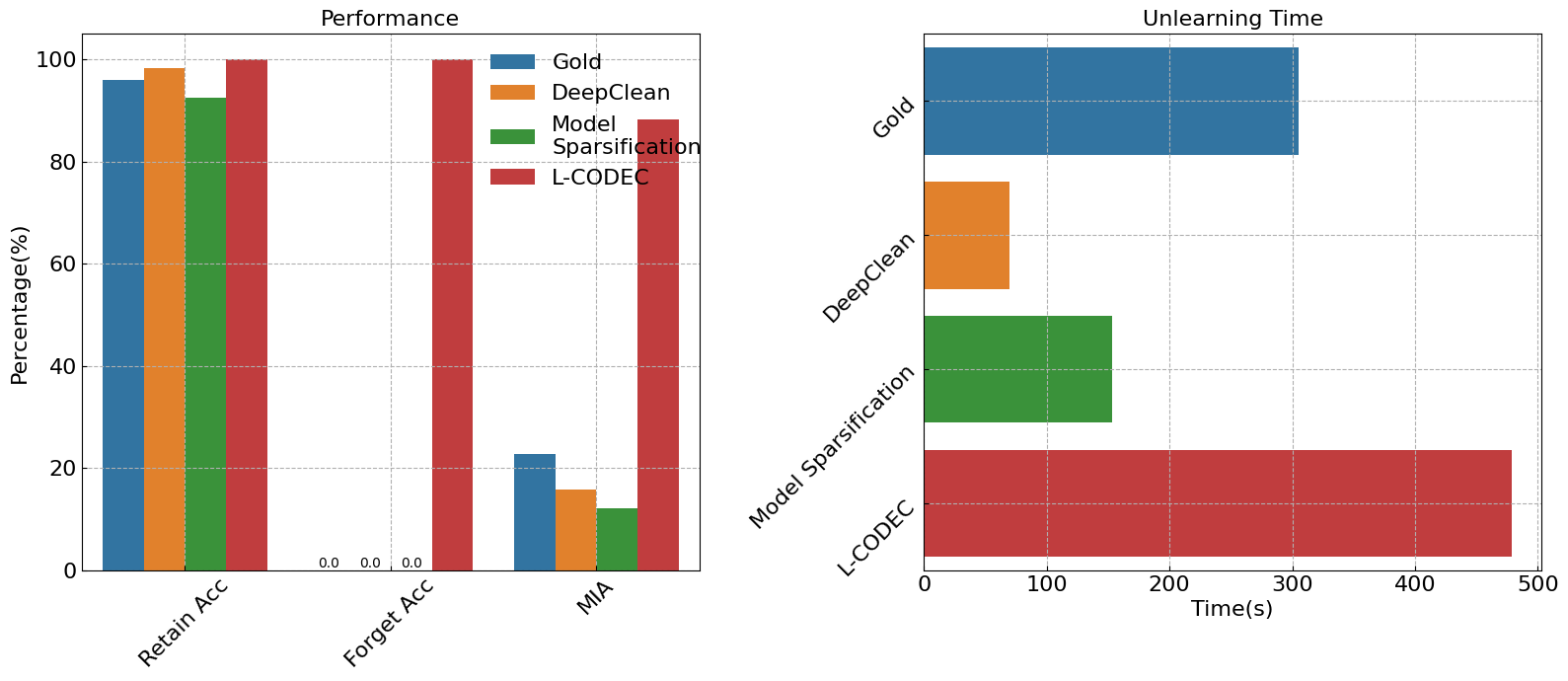}
\caption{Unlearning quality comparison between DeepClean and competitive unlearning algorithms (see Sec.~\ref{sec:experiments}). Accuracy and MIA should close to the Gold model. Unlearning time should be short.}
\label{fig:overview-results}
\end{figure}

We evaluate DeepClean on several standard image datasets and show that it can remove sensitive information without significantly impacting performance on other attributes. Experiments performed in this paper across different network architectures and datasets demonstrates an excellent performance of the proposed method reducing protected information from the original models while maintaining high performance on the retain dataset (Fig.~\ref{fig:overview-results}). To our knowledge, this represents the first application of the FIM diagonal to machine unlearning for computer vision models. Our results demonstrate that selectively forgetting information from neural networks can help address ethical concerns without requiring full model retraining.

\medskip\noindent\textbf{Contributions} 
\begin{itemize} 

\item A machine unlearning method for selectively removing sensitive knowledge encoded in trained models is developed and experimentally validated. This is accomplished by computing the diagonal of Fisher Information Matrix (FIM) based on the forget set compared to the FIM diagonal computed on the retain dataset. The rationale is to mark those weights that contribute to memorizing the forget set more than they do for the retain set. Once we identify them, we can tune them to decrease the information about the forget set while keeping the performance on the retain set. 

\item A set of practical rules that a good machine unlearning algorithm should satisfy is proposed.

\item The method is lightweight, intuitive, and can handle the unlearning task for a variety of model types including CNNs-based models with a rigorous machine unlearning definition. 

\item The developed method is efficiently capable to identify weights to be retrained in order to achieve the desired unlearning objective without the requirement to maintain weights or gradient information, in contrast to other algorithms that do the unlearning process by tracking weights and gradient information during the training phase.

\end{itemize}

\section{Related work}
\label{sec:related_work}

Machine unlearning was first introduced by Cao and Yang in 2015 with the goal to make machine learning systems forget~\cite{cao2015towards}.  Increasing interest due to privacy regulations (‘\emph{right to be forgotten}’) and potential use for error and bias removal has led to more studies, many that are model agnostic~\cite{Xu_2023,shaik2023exploring,xu2023machine,mercuri2022introduction,nguyen2022survey}. Methods for unlearning in a trained model are typically either \emph{exact} or \emph{approximate}, differentiated by completely or partially removing the influence of specific data points. Methods have been proposed for unlearning over data sample, class, feature, sequence, and graph~\cite{nguyen2022survey,xu2023machine}. A straightforward and exact method of unlearning is to re-train the model without the sensitive data in the training set. However, this is impractical for many use-cases where the trained model has required significant expense to learn, potentially millions of dollars for the largest models today. Alternatively, a model could be fine-tuned on data to be retained, however this can lead to catastrophic forgetting~\cite{2017PNAS..114.3521K} of unseen data points and may incompletely remove information that should be forgotten. More sophisticated exact approaches have been proposed~\cite{ijcai2022p556,Thudi2021OnTN}. In this paper, we contribute new methods for the approximate category of unlearning that makes limited parameter updates to the model to approximate exact unlearning. Approximate methods are generally more cost effective and efficient, scaling to large models. Within approximate methods, approaches can be divided into a variety of data reorganization or model manipulation techniques.  

The DeepClean approach presented here is an improved model-agnostic approximate method in the category of model manipulation, supporting sample and class requests, and is computationally efficient.  Prior work in approximate machine unlearning for model manipulation includes the use of influence functions~\cite{pmlr-v70-koh17a,guo2022efficient} and approximation of the Fisher Information Matrix~\cite{pmlr-v119-wu20b}.

Influence functions have emerged as one popular approach for machine unlearning. Proposed by Koh and Liang in 2017~\cite{pmlr-v70-koh17a}, this method leverages influence functions to identify individual training examples that have an outsized influence on a model's predictions. Recent researches conducted by \cite{golatkar2021mixed}, \cite{jia2024model} and \cite{mehta2022deep} show that by removing or downweighting these influential points, the model can ``forget'' specific attributes without requiring full retraining. A key advantage of influence functions is providing interpretability - the examples identified make clear what specific information is being removed from the model. However, estimating influence across all training data can be computationally expensive.

An alternative is to approximate the Fisher Information Matrix, which captures the sensitivity of model parameters to changes in the training data. \cite{pmlr-v119-wu20b} demonstrated that approximating and inverting the FIM makes it possible to selectively forget parts of the training set in a principled way, without having to retrain from scratch. Though more efficient, the FIM approach lacks the example-level interpretability of the influence function method.  An advantage for the FIM is that unlearning random samples and labels can be done simultaneously.  In comparison to other approaches using FIM, our DeepClean method 
FisherMask is another recent work \cite{liu2023unlearning} uses the approximate FM to mask model parameters most relevant to the forget set.

Differential privacy is strongly related to the unlearning goal, seeking to prevent information leaking through the output of models or functions \cite{dwork2014algorithmic}. \cite{cummings2018role} highlight the necessity of training machine learning models in a way that respects differential privacy in order to adhere to GDPR regulations. However, differential privacy does not allow data samples that have been used in model training to be forgotten.

In the experimental section we show DeepClean improves over recent work by \cite{golatkar2021mixed}, a scrubbing procedure that removes information from the trained weights, 
RandomLabel~\cite{graves2021amnesiac}, in which the sensitive data is relabeled with incorrect labels, and Teacher~\cite{chundawat2023a}, which coordinate two models for retaining and forgetting respectively while minimizing KL Divergence.

\section{Proposed Method}\label{sec:proposed-method}
Our method, DeepClean, being conceptually simple and empirically effective, proceeds in two steps:
\vspace{0.1in}

\begin{enumerate}
    \item Identify which model weights are most responsible for remembering the set to forget, $D_f$, and \label{it:step-one}
    \item  Freeze all of the model except those weights, then fine-tune them on the set to remember, $D_r$, to reinforce the performance of that set.\\ \label{it:step-two}
\end{enumerate}

\noindent To find those weights, we will use the empirical FIM. Suppose $p(y|x, w)$ is the distribution of $y$, \eg, a data point's class, given weights $w\in \mathbb{R}^n$ and features $x\in \mathbb{R}^d$. For example, $p(y|x,w)$ could be the final output of a neural network with a softmax layer. Then the empirical FIM of$w$ given a dataset $D$ is: 
\begin{equation}
    I_D(w)=\frac{1}{|D|}\sum_{(y,x)\in D}\nabla_{w}\log p(y|x,w)\nabla_{w}\log p(y|x,w)^{T}.
    \label{eq:FIM}
\end{equation}
For $n$ weights, $I(w)$ is an $n{\times}n$ matrix, making computation expensive. A common approximation is to compute only the diagonal elements of $I_D(w)$ and set the rest to zero. If the $i$-th diagonal element is large, then $D$ contains a lot of information about the $i$-th parameter $w_i$. This is the approximation we use in this work. 

We can use the FIM to identify the weights for Step~\ref{it:step-one}. First we define
\begin{equation}
r(w_i):=\frac{I_{D_f}(w_i)}{I_{D_r}(w_i)},
    \label{eq:ratio-forget-retain}
\end{equation}
where $I_{D}(w_i):=(I_{D}(w))_{i,i}$, the $i$th diagonal element of the empirical FIM. This ratio captures whether there is more information about $w_i$ in $D_f$ or $D_r$. 

For Step~\ref{it:step-two}, we first pick a threshold $\gamma$. With this, we can select the weights that are most informed by $D_f$ with the rule $r(w) > \gamma$, ~\ie, we have the sets $W_r = \{w_i: r(w_i)\leq \gamma\}$ and $W_f = \{w_i: r(w_i)>\gamma\}$. We then proceed to freeze all weights $W_r$ and fine-tune the weights $W_f$ on the retain set $D_r$ with initialization $W_f=0$. This reduces $r(w_i)$ for all $w_i\in W_f$ without requiring updating all of the model, so the method is especially advantageous when $|W_r|$ is larger than $|W_f|$.

\section{Experimental Evaluation}\label{sec:experiments}
\label{sec:results}

{\flushleft{\textbf{Datasets and Models}}}\quad
DeepClean is evaluated on image classification using MNIST dataset~\cite{deng2012mnist} that has $60,000$ training images and $10,000$ testing images with size $24\times 24$; Cifar-10 and Cifar-100~\cite{krizhevsky2009learning} both have $50,000$ training images and $10,000$ for testing with size $32\times 32$. 
Like many unlearning methods~\cite{Lin_2023_CVPR, tarun2023deep}, we use ResNet18~\cite{he2015deep}, and VGG-16~\cite{DBLP:journals/corr/SimonyanZ14a} to test the applicability of DeepClean. All experiments are performed on AWS g5.8xlarge EC2 instance with Intel(R) Xeon processors and NVIDIA(R) A10 (24GB) graphic card comes with Python3.8, PyTorch v2.0.1, torchvision v0.15.2 and CUDA v11.7. All models are trained with Stochastic Gradient Descent optimizer with a momentum of 0.9, weight decay of 5e-4, and an initial learning rate of 0.1, but different learning rate schedules. For Cifar-100, 200 epochs are used with a multi-step learning rate scheduler and learning rate is reduced by a factor of 5 at epoch 60, 120 and 160. For Cifar-10, models are trained with 20 epochs with learning rate decreased at epoch 6, 12 and 16. For MNIST, 10 epochs are used and learning rate is decreased at epoch 4 and 8. Random data augmentation is applied in the training to avoid overfitting.

{\flushleft{\textbf{Unlearning Tasks}}}\quad
We consider two unlearning scenarios: (i) random sample unlearning(RN), as in Golatkar et al.~\cite{golatkar2021mixed}, and (ii) entire class unlearning(Label) as in Tarun et al.~\cite{tarun2023fast}. For scenario (i) we unlearn randomly selected samples comprising \textit{10\%} of the total. For scenario (ii) we unlearn \textit{label 0}, which represents the class \textit{Number 0}, \textit{Airplanes} and \textit{Aquatic mammals} for the three respective datasets.

{\flushleft{\textbf{Evaluating Unlearning Algorithms}}}\quad
A good unlearning algorithm should balance task performance across:
    \begin{itemize}
        \item \textbf{Utility:} The unlearned model's capability to perform the original task, that the pre-trained model was designed for, on the retain set $\textit{D}_\textit{r}$. 
    \end{itemize}
    \begin{itemize}
        \item \textbf{Unlearning Quality:} How well the unlearning algorithm is able to remove information involved in ${\textit{D}_\textit{f}}$ from the pre-trained model.
    \end{itemize}
    \begin{itemize}
        \item \textbf{Efficiency:} The feasibility of the unlearning algorithm to run on scale and achieve a good performance in terms of Utility and Unlearning Quality, explained above, with reduced execution time, compute and storage.
    \end{itemize}
Several common metrics are widely used in evaluating unlearning algorithms. For the utility task, classification accuracy for ${\textit{D}_\textit{r}}$ is widely used, with higher ${\textit{Acc}_{\textit{D}_\textit{r}}}$ being better. For efficiency, unlearning time is a good measure, with shorter times being better. For the unlearning quality, classification accuracy for ${\textit{D}_\textit{r}}$ and Membership Inference Attack (MIA)-- the probability of an attacker successfully determine whether a particular data record was part of the training set, first introduced by \cite{homer2008resolving} in 2008 and formalized by \cite{dwork2015robust} in 2015, are common metrics. As Chundawat et al.~\cite{chundawat2023a} pointed out, ${\textit{Unlearned Acc}_{D_f}}{=}0$ and ${\textit{Unlearned MIA}{=}0}$ are not the optimal results. They discussed the \textit{Streisand effect}, where the unlearned model deliberately gives incorrect prediction to reduce ${\textit{Acc}_{\textit{D}_\textit{f}}}$ and \textit{MIA}, which can still lead to information leakage. 
\\

\fbox{\begin{minipage}{32em}
    A good unlearning algorithm should produce a model that closely matches the performance of one trained on only $D_r$, which we will call the \textit{gold model} (this definition is from \cite{chundawat2023a}).
\end{minipage}}
\\

We use this definition as the foundation for examining unlearning quality, where  we believe a good unlearning should not be wrongly predict ${\textit{D}_\textit{f}}$ but treat them as never seen before. Furthermore, we argue that for the unlearning scenario (i), ``random samples unlearning'', if $\textit{Gold Acc}_{\textit{D}_\textit{f}}$ is significantly greater than $ \textit{Gold Acc}_{\textit{D}_\textit{test}}$, the unlearned $ \textit{Acc}_{\textit{D}_\textit{f}}$ should be smaller than $\textit{Gold Acc}_{\textit{D}_\textit{f}}$ and close to $ \textit{Gold Acc}_{\textit{D}_\textit{test}}$. From the data perspective, the \textit{gold model} that has never seen ${\textit{D}_\textit{f}}$ should treat those samples like $\textit{D}_\textit{test}$. The violation of this principle indicates potential information leakage from ${\textit{D}_\textit{r}}$ to ${\textit{D}_\textit{f}}$ during the training of the \textit{gold model}.

{\flushleft{\textbf{Evaluation Metrics}}}\quad Based on the definition of good, we use the following metrics to evaluate the effectiveness of unlearning:
    \begin{itemize}
        \item ${\textit{Acc}_{\textit{D}_\textit{r}}}$ is the unlearned model's classification accuracy on the retain set.
    \end{itemize}
    \begin{itemize}
        \item ${\Delta\textit{Acc}_{\textit{D}_\textit{f}}}$ = ${\mathrm{Unlearned~}\textit{Acc}_{\textit{D}_\textit{f}}} - {\mathrm{Gold~}\textit{Acc}_{\textit{D}_\textit{f}}}$, measures the unlearned model's classification deviation from the \textit{gold model}.
    \end{itemize}
    \begin{itemize}
        \item ${\Delta \textit{MIA}}$ = ${\mathrm{Unlearned~}\textit{MIA} - {\mathrm{Gold~}\textit{MIA}}}$, measures the \textit{MIA} deviation from the \textit{gold model}. We adopted a linear logistic regression implementation from~\cite{tarun2023deep}.
    \end{itemize}
    \begin{itemize}
        \item \textit{Unlearning time} is the time used to perform the unlearning.
    \end{itemize}

{\flushleft{\textbf{Implementation Details}}}\quad
During the unlearning stage: The loss function used is a Cross Entropy Loss. Pre-trained models are unlearned on top of ${\textit{D}_\textit{r}}$ without any augmentation. We run this for 3 epochs for Cifar-10 and Cifar-100 and 1 epoch for MNIST with Adam optimizer~\cite{kingma2014adam}. The algorithm uses cosine annealing learning rate scheduler with initial learning rate $1\mathrm{e}{-3}$. A hyperparameter introduced by DeepClean is the $\gamma$ used to determine ${\textit{D}_\textit{f}}$ important weights ${\textit{W}_{\textit{D}_\textit{f}}}$. After conducting initial experiments, we observe that $\gamma{=}2$ works well enough for unlearning scenario (ii). For unlearning scenario (i), $\gamma{=}1.1$ works well for most experiments except unlearning VGG-16 with CIFAR-100. The $\gamma$ selection does have an impact in terms of utility and unlearning quality as discussed in our ablation studies.

\begin{table}[ht]
\centering
\small 
\begin{tabular}{P{1.5cm}P{2cm}P{1.5cm}P{1.5cm}P{1.5cm}P{1.5cm}P{1.5cm}}
\multicolumn{2}{c}{\multirow{2}{*}{}} & \multicolumn{2}{c}{VGG-16} & \multicolumn{2}{c}{ResNet-18} \\
 &  & RN & Label & RN & Label \\
 \midrule
\multirow{4}{*}{Gold} & ${\textit{Acc}_{\textit{D}_\textit{r}}}\%$ &92.79  &91.78  &94.50  &95.95 \\

 & ${\textit{Acc}_{\textit{D}_\textit{f}}}\%$ &92.60  &0.00 &94.60  &0.00  \\
 & ${\textit{MIA}}\%$ &73.32  &24.68  &74.71  &22.84  \\
 & Time s &201  &228  &284  &305 \\
\midrule
\multirow{4}{*}{DeepClean} & ${\textit{Acc}_{\textit{D}_\textit{r}}}\%$ &95.62  &90.92  &98.74  &98.27 \\

 & ${\Delta\textit{Acc}_{\textit{D}_\textit{f}}}\%$ &\textit{\textbf{-3.88}}  &\textit{\textbf{+0.00}}  &\textit{\textbf{-3.68}}  &\textit{\textbf{+0.00}}  \\
 & ${\Delta\textit{MIA}}\%$ &\textit{\textbf{-2.84}}  &+3.64  &\textit{\textbf{-5.33}}  &\textit{\textbf{-8.04}}  \\
 & Time s &\textit{\textbf{60}}  &\textit{\textbf{60}}  &\textit{\textbf{71}}  &\textit{\textbf{70}}  \\
\midrule
\multirow{4}{*}{Sparse MU} & ${\textit{Acc}_{\textit{D}_\textit{r}}}\%$ &88.81  &89.87  &91.89  &92.57  \\
 & ${\Delta\textit{Acc}_{\textit{D}_\textit{f}}}\%$ &-10.54  &+0.00  &-8.32  &+0.00  \\
 & ${\Delta\textit{MIA}}\%$ &-9.88  &\textit{\textbf{+1.84}}  &-12.82  &-10.16  \\
 & Time s &133  &133  &153  &153  \\
\midrule
\multirow{4}{*}{L-CODEC} & ${\textit{Acc}_{\textit{D}_\textit{r}}}\%$ &\textit{\textbf{99.85}}  &\textit{\textbf{99.85}}  & \textit{\textbf{100.00}} & \textit{\textbf{100.00}} \\
 & ${\Delta\textit{Acc}_{\textit{D}_\textit{f}}}\%$ & +7.3  & +99.96  &+5.40  & +100.00 \\
 & ${\Delta\textit{MIA}}\%$ & +17.54 & +67.96 &+14.17  & +65.44 \\
 & Time s & 245  & 276 & 366 & 479 \\
 \midrule
\multirow{4}{*}{\parbox{1.5cm}{Fisher\\ Forgetting}} & ${\textit{Acc}_{\textit{D}_\textit{r}}}\%$ &10.00  &11.10  & 9.97 & 12.13 \\
 & ${\Delta\textit{Acc}_{\textit{D}_\textit{f}}}\%$ & -84.50 & +0.00 &-82.82  & +0.00 \\
 & ${\Delta\textit{MIA}}\%$ &+22.25 & +76.16 &-70.70  & +55.22 \\
 & Time s & 3743  & 3792 & 3260 & 3193 \\
\end{tabular}

\caption{DeepClean compared with the other 3 influence function based algorithms, tested on Cifar-10 for both unlearning scenarios. Deepclean outperforms the others in most evaluation metrics. The ${\textit{l}_1 \ sparse \ MU + linear \ decaying \ \gamma}$ achieves similar forgetting performance but worse utility performance. L-CODEC performs poorly across all unlearning tasks, and we did notice that only less than 50k weights are updated with small magnitude changes, which leads to this poor unlearning performance.}
\label{tab:headline table}
\end{table}

{\flushleft{\textbf{Baselines Used}}}\quad
We compare DeepClean with few unlearning algorithms that able to perform both unlearning scenarios, a baseline model and the gold model.
    \begin{itemize}
        \item \textbf{Retrain from scratch (Gold):} Retrained model from scratch with retain set ${\textit{D}_\textit{r}}$
    \end{itemize}
    \begin{itemize}
        \item \textbf{Fine-tune}: Further fine-tune the pre-trained model for few more epochs on ${\textit{D}_\textit{r}}$ with same learning rate as the DeepClean.
    \end{itemize}
    \begin{itemize}
        \item \textbf{Model sparsification (Sparse MU):} Jinghan and Jiancheng demonstrated that model sparsification by pruning out weights can be an effective unlearning method~\cite{jia2024model}. We use the best set up they experimented with: ${\textit{l}_1 sparse \ MU}$\\{\textit{$+ linear \ decaying \ \gamma$}}.
    \end{itemize}
    \begin{itemize}
        \item \textbf{Fisher Forgetting:} a corrective Newton step is applied utilizing the Fisher Information Matrix. Fisher noise is introduced, drawn from a Gaussian distribution, to the model's weights to eliminate the information of ${D_f}$~\cite{golatkar2020eternal}.
    \end{itemize}
    \begin{itemize}
        \item \textbf{L-CODEC:} Like the Fisher forgetting, \cite{mehta2022deep} uses a variant of a new conditional independence coefficient to identify a subset of the model parameters to perform the unlearning task. We adopt the best set of hyperparameter found in their experiments. This algorithm requires to keep intermediate gradient information during the model training, which contradicts with our unlearning definition(see the middle of \cref{sec:intro}). Considering it is an influence function based method, it is added in the comparison.
    \end{itemize}
    \begin{itemize}
        \item \textbf{Teacher:} A student-teacher framework with an incompetent and competent teacher trained with ${\textit{D}_\textit{r}}$, where the incompetent teacher model is utilized to perform unlearning and the competent teacher model is used to maintain information~\cite{chundawat2023a}. We use the same learning rate $1\mathrm{e}{-3}$ trained for 10 epochs for Cifar-10 and Cifar-100, 5 epochs for MNIST.
    \end{itemize}

\begin{table*}[hb]
\centering
\fontsize{8}{11}\selectfont
\begin{tabular}{c c c c c c c}
Datasets & Models & Unlearning Algorithm & ${\textit{Acc}_{\textit{D}_\textit{r}}}\%$ & ${\Delta\textit{Acc}_{\textit{D}_\textit{f}}}\%$ & $\Delta \textit{MIA}\%$ & Time (s)\\
\hline
&  & Fine-tune & \textbf{\textit{99.85}} & +0.66 & +0.31 & 71\\
& ResNet18 & DeepClean & 98.74 & \textbf{\textit{-3.68}} & \textbf{\textit{-5.33}} & 71\\
&  & Teacher & 65.74 & -74.14& -74.71 & 110\\
Cifar-10\\ \cline{2-7}
&  & Fine-tune & \textbf{\textit{98.47}} & +0.18& +3.94 & 54\\
& VGG-16 & DeepClean & 95.62 & \textbf{\textit{-3.88}} & \textbf{\textit{-2.84}} & 60\\
&  & Teacher & 69.19 & -61.08& -74.71 & 80\\
\hline
&  & Gold & 100.00 & 61.00 & 13.78 & 2806\\
&  & Fine-tune & \textbf{\textit{99.98}} & +38.94 & +77.52 & 71\\
& ResNet18 & DeepClean & 99.87 & \textbf{\textit{+26.88}} & +40.60 & 69\\
&  & Teacher & 44.64 & -52.74 & \textbf{\textit{-11.78}} & 111\\
&  & Fisher Forgetting & 1.05 & -60.22 & -13.78 & 40820\\
Cifar-100 \\ \cline{2-7}
&  & Gold & 100 & 62.53 & 30.90 & 1832\\
&  & Fine-tune & \textbf{\textit{99.97}} & +37.15 & +60.42 & 52\\
& VGG-16 & DeepClean & 88.69 & \textbf{\textit{+3.85}} & \textbf{\textit{+23.06}} & 60\\
&  & Teacher & 34.01 & -50.81 & -30.90 & 84\\
&  & Fisher Forgetting & 1.05 & -61.51 & -30.88 & 29861\\

\end{tabular}
\caption{Unlearning scenario (i) performance evaluation for Cifar-10 and Cifar-100 with ResNet18 and VGG-16 for more datasets and baselines. Smaller deviations from the Gold model are better. DeepClean still outperforms other unlearning algorithms in all utility, unlearning tasks and efficiency.}
\label{tab:Random unlearn}
\end{table*}

{\flushleft{\textbf{Comparison with Baselines}}}\quad

\subsection*{\normalsize \textbf{Unlearning Scenario (i)}}
In our initial experiments, we focus on unlearning scenario (i), termed as random sample unlearning (RN), without incorporating any data augmentation. As demonstrated in \cref{tab:headline table}, DeepClean consistently surpasses other influence function-based unlearning algorithms across most evaluation metrics when tested with Cifar-10 data. The only exception is model sparsification-based unlearning, which exhibits performance comparable to DeepClean, albeit with inferior results across all utility, unlearning, and efficiency measures.

We observe that both L-CODEC and Fisher Forgetting perform suboptimally on the unlearning task. This suggests that the effectiveness of unlearning by introducing noise to the model weights, identified as significant using an approximation of the full Hessian, is limited. Both algorithm yield a high $\Delta \textit{MIA}$, albeit for different reasons. L-CODEC's high ${\textit{Acc}_{\textit{D}_\textit{f}}}$ indicates that too few weights are updated. Conversely, Fisher Forgetting updates an excessive number of weights, leading to near random model predictions.

This bias in the number of updated weights not only results in a deviation of the model weights from the gold model but also alters the output distribution. This makes an MIA more likely to occur, ultimately leading to a high $\Delta \textit{MIA}$. These findings highlight the limitations of these two influence function-based algorithms in real-world unlearning and utility scenarios.

In the subsequent experiments presented in \cref{tab:Random unlearn}, further fine-tuning consistently delivers the best utility performance, as anticipated, but at the expense of an increased ${\textit{Acc}_{\textit{D}_\textit{f}}}$. Overall, DeepClean continues to outperform other unlearning algorithms, demonstrating balanced performance across utility, unlearning, and efficiency tasks.

\subsection*{\normalsize \textbf{Unlearning Scenario (ii)}}
In our subsequent experiments, we explore unlearning scenario (ii), termed as label unlearning (Label), without incorporating any data augmentation.We observe performance patterns similar to scenario (i) as shown in \cref{tab:headline table} Label columns. Unlearning through model sparsification closely mirrors DeepClean in the unlearning task but falls short in utility and efficiency tasks. L-CODEC excels in the utility task, albeit at the expense of not unlearning any $\textit{D}_\textit{f}$. Fisher Forgetting's propensity to remove excessive information from the model aids in unlearning $\textit{D}_\textit{f}$ but compromises the utility task. Additional experiments with other unlearning algorithms and datasets are presented in \cref{tab:Label 0 unlearn}, where DeepClean continues to outperform across most metrics.

Across all conducted experiments, unlearning through model sparsification shows potential, yet it consistently trails behind in utility, unlearning, and efficiency measures, further underscoring the robustness of DeepClean. The Hessian-based methods, L-CODEC and Fisher Forgetting, exhibit limitations in their effectiveness across both unlearning scenarios. L-CODEC's strategy of updating fewer weights and Fisher Forgetting's approach of excessive weight updates have led to suboptimal performance, increased susceptibility to MIAs, and poor utility task performance. The Teacher method demonstrates high performance variation across different datasets and model architectures.

In conclusion, our findings from both unlearning scenarios highlight the efficacy of DeepClean as the most proficient algorithm for unlearning tasks. Its consistent performance across different models and datasets, coupled with its ability to maintain a balance between utility, unlearning, and efficiency tasks, also importantly, the adherence to the practical unlearning definition we defined in \cref{sec:intro} makes it an ideal choice for practical unlearning and utility scenarios.


\begin{table*}[h]
\centering
\fontsize{8}{11}\selectfont
\begin{tabular}{c c c c c c c}
Datasets & Models & Unlearning Algorithm & ${\textit{Acc}_{\textit{D}_\textit{r}}}\%$ & ${\Delta\textit{Acc}_{\textit{D}_\textit{f}}}\%$ & $\Delta\textit{MIA}\%$ & Time (s)\\
\hline
&  & Gold & 99.88 & 0.00 & 1.00 & 88\\
&  & Fine-tune & \textbf{\textit{99.98}} & +99.93 & +80.70 & 11\\
& ResNet18 & DeepClean & 99.95 & \textbf{\textit{+0.00}} & \textbf{\textit{-0.63}} & 22\\
&  & Teacher & 97.81 & +21.49 & +97.89 & 40\\
&  & Fisher Forgetting & 8.77 & +39.07 & +96.43 & 4427\\
MNIST \\ \cline{2-7}
&  & Gold & 99.91 & 0.00 & 1.42 & 219\\
&  & Fine-tune & \textbf{\textit{99.90}} & +99.98 & +96.44 & 21\\
& VGG-16 & DeepClean & 99.86 & \textbf{\textit{+0.00}} & \textbf{\textit{+2.26}} & 49\\
&  & Teacher & 98.75 & +45.00 & +98.58 & 88\\
&  & Fisher Forgetting & 12.82 & +0.00 & +93.36 & 3889\\
\hline
&  & Fine-tune & \textbf{\textit{99.68}} & +23.00 & -13.97 & 71\\
& ResNet18 & DeepClean & 98.27 & \textbf{\textit{+0.00}} & \textbf{\textit{-8.04}} & 70\\
&  & Teacher & 64.49 & +16.16 & +12.48 & 111\\
Cifar-10 \\ \cline{2-7}
&  & Fine-tune & \textbf{\textit{98.50}} & +14.06 & -8.16 & 52\\
& VGG-16 & DeepClean & 90.92 & \textbf{\textit{+0.00}} & +3.64 & 60\\
&  & Teacher & 68.34 & +24.54 & \textbf{\textit{-0.14}} & 80\\
\hline
&  & Gold & 99.99 & 0.00 & 28.16 & 3038\\
&  & Fine-tune & \textbf{\textit{99.98}} & +99.80 & +30.24 & 78\\
& ResNet18 & DeepClean &\textbf{\textit{99.98}} & \textbf{\textit{+0.00}} & \textbf{\textit{-24.96}} & 75\\
&  & Teacher & 44.00 & +2.60 & -28.16 & 85\\
&  & Fisher Forgetting & 1.04 & +0.00 & +71.84 & 40048\\
Cifar-100 \\ \cline{2-7}
&  & Gold & 84.26 & 0.00 & 62.6 & 2520\\
&  & Fine-tune & \textbf{\textit{99.97}} & +99.60 & \textbf{\textit{+4.80}} & 58\\
& VGG-16 & DeepClean & 96.83 & \textbf{\textit{+0.00}} & -45.2 & 63\\
&  & Teacher & 35.36 & +3.40 & -62.6 & 61\\
&  & Fisher Forgetting & 0.99 & +0.86 & -61.63 & 33022\\

\end{tabular}
\caption{Unlearning scenario (ii) performance evaluation for MNIST, Cifar-10, and Cifar-100 with ResNet18 and VGG-16. We use 1 epoch for MNIST fine-tuning considering the complexity of the dataset is not comparable with Cifar-10 and Cifar-100.}
\label{tab:Label 0 unlearn}
\end{table*}

{\flushleft{\textbf{Ablation studies}}}\quad
We run two sets of ablation studies for DeepClean.

\begin{table*}
\fontsize{8}{11}\selectfont
\centering
\begin{tabular}{c c c c c c }
Datasets & Unlearning Scenario & Unlearning Algorithm & ${\textit{Acc}_{\textit{D}_\textit{r}}}\%$ & ${\Delta\textit{Acc}_{\textit{D}_\textit{f}}}\%$ & $\Delta\textit{MIA}\%$\\
\hline
&  & Gold & 92.79 & 92.60 & 73.32\\
& RN & Zero Weights & 16.50 & -75.98 & -73.32\\
& & DeepClean & \textbf{\textit{95.62}} & \textbf{\textit{-3.88}}& \textbf{\textit{-2.84}}\\
Cifar-10 \\ \cline{2-6}
&  & Gold & 91.78 & 0.00 & 24.68\\
& Label & Zero Weights & 12.78 & +0.00 & \textbf{\textit{+1.14}}\\
& & DeepClean & \textbf{\textit{90.92}} & \textbf{\textit{+0.00}} & +3.64\\
\hline
&  & Gold & 100 & 62.53 & 30.90\\
& RN & Zero Weights & 1.39 & -61.31 & \textbf{\textit{+15.80}}\\
& & DeepClean & \textbf{\textit{88.69}} & \textbf{\textit{+3.85}} & +23.06\\
Cifar-100 \\ \cline{2-6}
&  & Gold & 84.26 & 0.00 & 62.60 \\
& Label & Zero Weights & \textbf{\textit{99.86}} & +0.00 & -62.58\\
& & DeepClean & 96.83 & \textbf{\textit{+0.00}} & \textbf{\textit{-45.2}}\\
\end{tabular}
\caption{Ablation study of the necessity of retraining after the ${\textit{D}_\textit{f}}$ important weights' info has been removed}
\label{tab:zero weights comparison}
\end{table*}

\subsection*{\normalsize \textbf{(i) Necessity of retraining}}
Intuitively, after determining the important weights for ${\textit{D}_\textit{f}}$, a simple mask of these weights would remove the information that contribute to predictions. However, because the model's weights are collectively optimized in the first place, such direct removal would not only remove the ${\textit{D}_\textit{f}}$ information but also harm information about ${\textit{D}_\textit{r}}$. We argue that a few more epochs of retraining is necessary to align the weights better. In \cref{tab:zero weights comparison}, we compare the Zero Weights Initialization (Zero Weights) with DeepClean using VGG-16 for both unlearning scenarios random samples unlearning (RN) and label unlearning (Label) on Cifar-10 and Cifar-100. We empirically show that retraining better aligns the unlearned model's performance with the \textit{gold model}.
\begin{figure}[ht]
\centering
\includegraphics[width=0.55\textwidth]{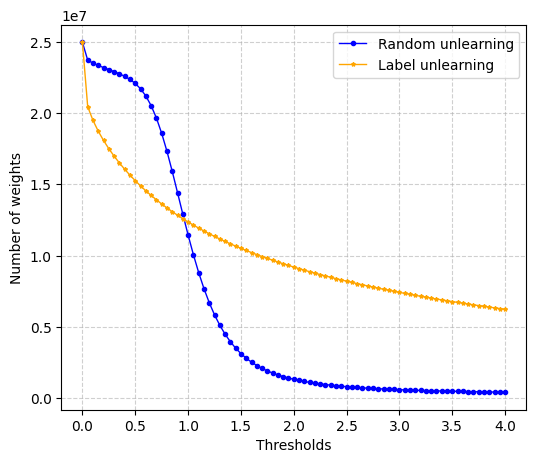}
\caption{Number of ${\textit{D}_\textit{r}}$ important weights vs. $\gamma$ for two unlearning scenarios on Cifar-10 and VGG-16. The threshold $\gamma$ controls how much of the model we will have to update to forget the influence of $D_f$. The range from $2$ to $3$ indicates potential sweet spots. Taking $\gamma$ close to $0$ leads to having to update most of the model. For both unlearning scenarios, $\gamma{=}2$ gives good $\Delta \textit{MIA}$ performance.}
\label{fig: weights vs thresholds plot}
\end{figure}

\subsection*{\normalsize \textbf{(ii) Threshold Sensitivity Analysis}}
\begin{figure*}[ht]
    \centering
    \begin{minipage}{0.45\textwidth}
        \centering
        \includegraphics[width=\textwidth]{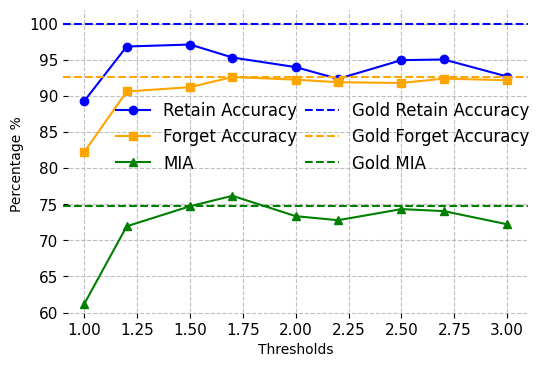} 
    \end{minipage}
    \begin{minipage}{0.45\textwidth}
        \centering
        \includegraphics[width=\textwidth]{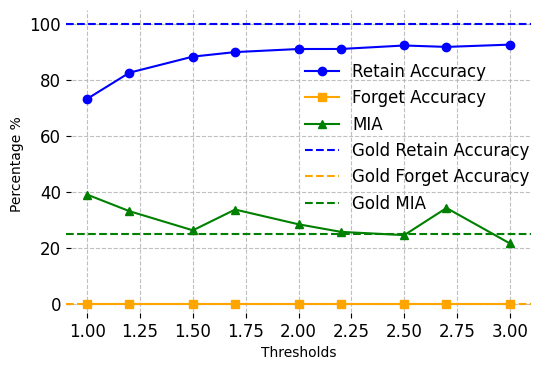} 
    \end{minipage}
    \caption{Utility and unlearning tasks' performance span across $\gamma$ range 2 to 3. The number of ${\textit{D}_\textit{r}}$ important weights decrease at an increasing pace within this range}
    \label{fig: forgetting performance vs. thresholds}
\end{figure*}

We introduce a hyper parameter $\gamma$ in DeepClean (see end of \cref{sec:proposed-method}), which has direct impact on the number of weights that will be fine-tuned. As shown in Fig.\ref{fig: weights vs thresholds plot}, $\gamma$=1 coincides with about half of the model's trainable weights for both unlearning scenarios. The selection of $\gamma$ should avoid $\gamma{<}1$ which would bring the algorithm closer to fully retrain a model. Choosing $\gamma{>}1$ gives better chance of balancing the trade-offs between efficiency, utility, and unlearning tasks. We run sensitivity analysis for both unlearning scenarios with VGG-16 on Cifar-10, with $\gamma{\in}\{1, 1.2, 1.5, 1.7, 2.0, 2.2, 2.5, 2.7, 3\}$ where weights numbers accelerate in converging. As \cref{fig: forgetting performance vs. thresholds} shows, for the selected range of thresholds, if evaluating the unlearning performance with $\Delta\textit{MIA}$, values on both sides of $\gamma{=}2$ are potential sweet spots for both unlearning scenarios. However,  based on the definition of good we discussed at the end of Evaluating Unlearning Algorithms section and the experiment results we observed where $\textit{Acc}_{\textit{D}_\textit{f}}$ potentially upward biased for Cifar-10, $\gamma$ close to 1 should be a better choice. Threshold values near $\gamma{=}2$ have \textit{MIA} close to \textit{Gold MIA}, but does not satisfy $\textit{Acc}_{\textit{D}_\textit{f}} < \textit{Gold Acc}_{\textit{D}_\textit{f}}$. Moving $\gamma$ towards 1 could mitigate the potential information leakage happened in \textit{gold model} retraining stage. Another benefit of selecting a smaller $\gamma$ for unlearning scenario (i) is with more weights being trained, utility performance is increased. Therefore, to balance utility task and unlearning task, and mitigate the issues discussed, we determined qualitatively for unlearning scenario (i) $\gamma{=}1.1$ is a good choice, and $\gamma{=}2$ for unlearning scenario (ii).

\section{Limitations}
\label{sec:discussion}
Our work on DeepClean has demonstrated its effectiveness in machine unlearning, providing a lightweight and flexible solution for removing sensitive information from trained models. However, there are several areas where further improvements and investigations could be made.

\medskip\noindent\textbf{Hyperparameter Selection}: The current approach to determining the threshold $\gamma$ is qualitative. While this approach has proven effective in our experiments, an automatic selection process based on meaningful and interpretable metrics could potentially improve the performance and generalizability of our method. This could transform the problem into a constrained optimization problem, where the optimal value of $\gamma$ is determined algorithmically based on the specific characteristics of the data and model.

\medskip\noindent \textbf{Pretrained Model Quality}: The effectiveness of an unlearning algorithm is inherently dependent on the quality of the start point, "pretrained" model, of the unlearning process. However, to the best of our knowledge, most of the unlearning studies focus on evaluating the "unlearned" model performance, with no attention paid to the "pretrained" model. We argue that for machine unlearning, common evaluation metrics like MIA and $\textit{Acc}_{\textit{D}_\textit{f}}$ are not sufficient. A metric that successfully measures the degree to which a model is saturated with training information could help in interpreting the unlearning variation for different models and datasets. Furthermore, such metrics could provide more information regarding the number and magnitude of model weights need to be updated for all influence-function based unlearning algorithms. For example, determining the hyperparameter $\gamma$ in DeepClean, and the amount of noise added in Fisher Forgetting.

\medskip\noindent\textbf{Dynamic Weight Updating}:  In the current implementation of DeepClean, the FIM diagonal elements are calculated once and the weights to be retrained are determined based on this calculation. However, it could be beneficial to recalculate the FIM diagonal elements during the fine-tuning stage, allowing for a joint optimization process. This would enable the weights to be updated dynamically, not only to give correct predictions but also to maximize the forgetting of the sensitive subset.

\section{Conclusions}
\label{sec:conclusion}
In this paper, we introduce DeepClean, a machine unlearning algorithm that approximates the FIM diagonal based on the retain and forget dataset. Our method efficiently identifies weights that require retraining without the need to track weights or gradient information. This characteristic makes our method lightweight and flexible, making it applicable not only to various CNN based models. Additionally, our approach is not limited to unlearning specific labels; it can also handle the unlearning of any subset of samples without multiple class labels. Our experimental results validate DeepClean’s efficacy in achieving its objectives while respecting a set of practical unlearning rules we believe are foundational. Furthermore, our ablation studies highlight the importance of retraining.

In summary, DeepClean emerges as a straight forward yet effective solution for resolving sensitive information from trained models, thereby addressing significant privacy concerns for both specific labels and also for random subsets of samples.

\par\vfill\par
\clearpage  

%
%
\bibliographystyle{splncs04}
\bibliography{main}
\end{document}